\documentclass[10pt, a4paper]{article}

\usepackage{lrec-coling2024} 
\usepackage{orcidlink}
\newcommand{\eg}{e.g.,\ }
\newcommand{\ie}{i.e.,\ }
\newcommand{\etal}{\textit{et al.\ }}

\title{Systemic Biases in Sign Language AI Research: \\A Deaf-Led Call to Reevaluate Research Agendas}

\name{Aashaka Desai \orcidlink{0000-0003-4601-8137},\ Maartje De Meulder \orcidlink{0000-0001-7607-5314},\ Julie A. Hochgesang \orcidlink{0000-0001-7450-2343},
     \\ \large \textbf{Annemarie Kocab \orcidlink{0000-0002-1888-9499}, and Alex X. Lu} \orcidlink{0000-0001-9568-3155}}

\address{University of Washington, University of Applied Sciences Utrecht/Heriot-Watt University,         \\Gallaudet University, Johns Hopkins University, Microsoft Research\\
        aashakad@cs.washington.edu, maartje.demeulder@hu.nl, julie.hochgesang@gallaudet.edu, 
        \\kocab@jhu.edu, lualex@microsoft.com}

\abstract{Growing research in sign language recognition, generation, and translation AI has been accompanied by calls for ethical development of such technologies. While these works are crucial to helping individual researchers do better, there is a notable lack of discussion of systemic biases or analysis of rhetoric that shape the research questions and methods in the field, especially as it remains dominated by hearing non-signing researchers. Therefore, we conduct a systematic review of 101 recent papers in sign language AI. Our analysis identifies significant biases in the current state of sign language AI research, including an overfocus on addressing perceived communication barriers, a lack of use of representative datasets, use of annotations lacking linguistic foundations, and development of methods that build on flawed models. We take the position that the field lacks meaningful input from Deaf stakeholders, and is instead driven by what decisions are the most convenient or perceived as important to hearing researchers. We end with a call to action: the field must make space for Deaf researchers to \textit{lead} the conversation in sign language AI. }

\begin{document}

\maketitleabstract

\section{Introduction}

Applications of machine learning (ML) and artificial intelligence (AI) to sign languages have exploded over the past few years.
As large-scale sign language datasets emerge, 
a growing number of works apply data-driven AI methods from computer vision and natural language processing to solve various problems including sign language recognition, translation, and generation \cite{bragg2021fate, yin2021including, borstell2023ableist}.

At the same time, the field has been shaped by systemic barriers causing the historical and present exclusion of Deaf\footnote{We use 'deaf' to refer to audiological status, and 'Deaf' to refer to cultural identities. While the field of Deaf Studies is moving away from the use of deaf vs. Deaf \cite{kusters2017innovations}, here we prefer a more explicit signposting of identity. While we aim to be precise, the miscible nature of identity means at times, our usage is interchangeable, but our intent is not to use terms as a means to exclude.} people from it \cite{angelini}.
This includes the ableism and audism that shapes perceptions of Deaf communities and signed languages, as well as larger trends in STEM education that exclude Deaf individuals from being involved in research about them.
Borstell \cite{borstell2023ableist} shows that as many as 12\% of papers in sign language computing contain basic ableist terms, double the incidence of linguistics. 

Towards more equitable research, previous work has identified major issues in papers, and issued recommendations on how to improve sign language AI research from multiple perspectives, including ethical considerations in datasets, linguistic considerations, and community engagement (\eg \cite{fox2023best, de2022challenges, bragg2021fate, de-meulder-2021-good}).
While these efforts are critical to addressing the ableism and audism that permeates the field, they generally focus on individual interventions encouraging authors to do better. 

In our work, we reasoned that the systemic impact of excluding Deaf researchers from sign language AI research may be more subtle, and that a critical interrogation is needed of the assumptions and rhetoric that shape the research questions and methods 
in the field. 
In principle, even if each individual paper and research project followed best practices in responsible (sign language) AI, the collective direction of the field may still be misaligned with the interests and perspectives of most Deaf stakeholders.
Collectively, what problems and aspects of signed languages are considered worth studying, and who decides such? 

In other emerging fields, critical literature reviews have been crucial in redirecting research (\cite{mack2021we, spiel2022adhd, froehlich2010design}).
Inspired by these works, we conducted a hybrid literature review and position paper analyzing over 100 papers in sign language AI.

Our analysis identifies systemic biases in the current state of sign language AI research. 
We show that the majority of papers are motivated by solving perceived communication barriers for Deaf individuals, use datasets that do not fully represent Deaf users, lack linguistic grounding, and build upon flawed models. 
From these results, we take the position that the field suffers from a lack of intentional inclusion of Deaf stakeholders. Lacking meaningful and ongoing input from Deaf stakeholders, the field is instead driven by what approaches and modeling decisions are the most convenient. We end with a call to action: the field must make space for Deaf researchers to \textit{lead} the conversation in sign language AI.

\section{Positionalities and lived experiences}

Our analysis and positions are shaped strongly by our identities and positionalities. We are a group of five researchers: we all identify as deaf, Deaf or hard-of-hearing (DHH). Two of us are white, three are Asian. Our interdisciplinary team spans a range of fields and research interests, including machine learning and computer vision, Deaf Studies and applied language studies, linguistics language documentation/corpora, phonetics/phonology, HCI and accessibility, psycholinguistics, language acquisition, developmental psychology, and cognitive science. We recognize that we come from positions of literacy and educational privilege, which may not be representative of Deaf communities. Our daily communication encompasses a blend of signed, written, and for some of us, spoken languages. Collectively, our linguistic repertoires include ASL, International Sign, NGT, VGT, KSL, English, Dutch, Gujarati and Hindi, along with other languages. Our experiences with assistive hearing technologies vary, with some of us having used hearing aids in the past while others continue to use them. We have varied lived experiences, but share the experience of growing up deaf or hard-of-hearing and going to mainstreamed schools for all or most of our education. 
Some of us grew up signing. 
For some of us, signing has been a part of our lives from an early age, while others began signing in their teenage years.

That all authors of this paper are DHH is intentional. Our aim from the outset was to approach this research from explicitly DHH positionalities and to bring different viewpoints. Since deaf people are the primary stakeholders in sign language technologies, we believed it essential to foster a space where DHH researchers could engage in open discussions about biases in ML applications to sign languages. The act of suggesting that hearing collaborators may be contributing to systemic bias seen in the field puts undue burden on DHH authors to carefully manage what they say. 
Because every member was DHH, we were able to openly discuss systemic bias and extend our discussion to not only include very clear instances of ableist works but also delve into the more subtle effects of ingrained biases in sign language AI research. Similar spaces created by other DHH scholars have generated insightful discussions of issues central to Deaf stakeholders \cite{kusters2017innovations, chua20221001, o2023deaf}.

\section{Methods}
\subsection{Corpus creation}

Sign language computation research lies at the intersection of Natural Language Processing, Computer Vision, and Human-Computer Interaction/Accessibility. As no dedicated venues centralize the majority of relevant work, we turned to arXiv, where computational researchers often share preprints of their work. We retrieved all papers containing the term ``sign language'' in CS field on arXiv, scoping our search to papers January 2021 to November 2023. This yielded 222 papers.

As our review focuses on sign language AI, we exclude works that exclusively study human factors. 
For papers in sign language AI, we focus on ``receptive'' sign language models, models that accept a recording or representation of sign language as input.
Although work that focuses on sign language generation or avatars is also interesting and contributes to language understanding, these methods are relatively less developed \cite{yin2021including}. We reasoned focusing on receptive models would provide more diversified design decisions for analysis while reducing the volume of papers.
We also exclude works that do not center sign language (\eg uses sign language to demonstrate how methods generalize). 
Since our work focuses on sign \textit{language}, we include work that focuses on fingerspelling only if they explore fingerspelling in the context of a longer sentence, or if the work (erroneously) claims fingerspelling to be a complete language system. 
We exclude reviews, theses, and non-English works. 

Three authors reviewed paper abstracts against our inclusion criteria. Initially, two authors were assigned to each abstract. If there was a disagreement, a third author broke the tie. After filtering through inclusion criteria, we had 137 papers.

A limitation of arXiv is that works have not necessarily been peer reviewed. We only include published works from 2021-2022 (excluding 26 papers). As 2023 arXiv papers might be currently undergoing review, we include all preprints from that year that match our inclusion criteria. This gave us a total of 111 works for our systematic analysis.

\subsection{Systematic Literature Review} 

We developed a codebook iteratively through discussion between authors (see appendix). 
We track the datasets used in each paper alongside inputs to models and outputs of the model (\ie labels). We also note any prior models that papers build on (\ie pretraining). We additionally read the abstracts and introduction to understand how the paper is motivated. 
Two annotators coded each paper, and disagreements were resolved by a third annotator. 

\section{Results and Discussion}

We excluded 10 papers from our initially compiled list on further review as we found they did not match inclusion criteria. Our review thus consisted of a total of 101 papers, 21 from 2021 (peer-reviewed), 29 from 2022 (peer-reviewed), 51 from 2023 (arXiv). Most of these works focused solely on sign language recognition or translation as their main task, with a few looking at additional tasks like segmentation, sign spotting, etc. 

Of the 101 papers in our review, we find that 60 work with continuous sign language datasets, 26 work with isolated sign language datasets, 3 with a combination of isolated and continuous sign language datasets, and 11 work with fingerspelling data. 
Most datasets used are publicly available. Seven works collect their own private dataset. 
Below we discuss themes from our systematic review.

\subsection{Papers are motivated by perceived communication barriers}

In our review, we find that 64 papers primarily motivate their work as addressing barriers in communication between deaf people and hearing society or spoken language resources. 
Navigating a hearing world and resulting communication barriers are undeniably a central component of the lived deaf experience. However, sign languages are not merely ``communication tools'' \cite{hu2023self},
they are full languages, with a long history of being recognized as such \cite{de2019legal}. When ML research focuses singularly on the role sign languages play in provisioning access, it overlooks the history and diverse lived experiences of Deaf people, and misses out on exciting avenues for research, as we discuss below. 

We find in most papers, the description of communication barriers encountered by deaf individuals either implies or directly establishes an inherent connection between sign language use and hearing ability. 
First, many papers claim that sign languages are the ``primary form'' \cite{walsh2023gloss} or ``natural means" \cite{varol2021read} of communication for deaf people. 
However, not all deaf individuals know and use a signed language, and the signing communities extends beyond those who identify as Deaf. 
Even as individuals should have the right to self-determine what communication modalities they use in what contexts, the systemic suppression of sign languages means that many deaf people are not given sign language as an option in the first place.
By presenting an oversimplified claim that ``deaf people use sign language'', authors fail to pay credence to this long-standing oppression (as well as movements seeking equal status for signed languages) that complicate this relationship \cite{murray2019education}.

Second, there is a frequent narrative in the papers that suggests the primary hurdle in communication between `deaf' and `hearing' people is the `lack of a shared language', with some papers claiming that deaf people largely lack fluency in written languages (\eg \textit{``the globe’s [430 million DHH people] largely do not benefit from modern language technologies''} \cite{wang2023active}). 
This framing diminishes the multilingual and multimodal capabilities of deaf people \cite{kusters2017beyond}.
Often, deaf and hearing people \textit{do} share a common language, but deaf people might not have physical access to auditory languages. Most sign languages do not have a commonly used written form and so deaf signers often learn to read and write in another language \cite{languages6020085}, even as some face (and overcome) barriers in acquisition of spoken languages.
Additionally, by fixating on how deaf people communicate exclusively, this framing portrays communication as one-sided when it is usually reciprocal and multimodal.
`Communication' for deaf people is much more complex than a mere translation between signed and spoken languages.

Third, perceived communication barriers are often used to argue that deaf people are not included into hearing society, and therefore experience adverse consequences.
For example, in their discussion of broader impact, Hu \etal \cite{hu2023signbert+} state that deaf people may ``feel isolated, lonely, or [have] other mental health issues when they face the communication barrier in daily life''. 
While it is true that inaccessibility impacts deaf people on a systemic and individual level, claims like these portray deaf people as deficient and in need of technological interventions (termed by \cite{morozov2013save} as `technosolutionism'), instead of more accurately recognizing that most deaf individuals already have developed strategies to navigate hearing society, and that any emerging technology will at least initially only be a small supplement to these strategies. 
Thus, this framing of deaf individuals is ideological, allowing authors to overstate the importance of their contributions to the daily lives of deaf people, at the expense of diminishing their existing repertoires. 

We note that not every paper that focuses on communication barriers frames poorly.
For example, Hossain at al. \cite{hossain2023edgcon} are careful to scope their claims to barriers in STEM education and design a method well aligned with the application.
However, we believe there are two distinct issues: first, in our reading, the majority of papers that do motivate their work as addressing communication barriers do have oversimplified or inaccurate views.
But second is the overall proportion of papers in the field that focus on mitigating communication barriers.

Addressing the second issue, in our view, this means the field disproportionately focuses on a single story: mitigating accessibility barriers, which is primarily understood to be ``deaf people's access to spoken language''. 
This means that receptive sign language models are mostly studied in the context of translation, overemphasizing the role of spoken language.
While this is an important issue, it is not the only framework in which sign language recognition can occur.
In our review, we find a few works that are motivated by exploration of sign language as a language in its own right, including models that annotate phonology \cite{tavella2022phonology}, or predict the iconicity of signs \cite{hossain2023edgcon}, but these are far less represented than translation works.
Sign languages \textit{are} different in many ways than spoken languages, and rather than considering these differences as inherent limitations that make building sign language technologies difficult, there is an opportunity to develop AI technologies that understand and center these differences to further our scientific understanding of the human capacity of language.
For example, as we further discuss in Section 4.3, most translation annotation schemes focus on flattening phonological differences between users to prioritize semantics, but differences in phonology can induce differences in meaning, as well as connect to the identity of the signer. Applications like these are currently underserved by sign language AI.


\subsection{Models use datasets misaligned with target users}

Across all papers, we identified 43 different publicly available sign language datasets. 16 datasets use solely DHH contributors, 3 datasets use solely interpreters, 11 datasets include a mix of contributors, and 12 datasets do not specify contributor qualifications. While this heterogeneity in dataset contributors seems promising at the surface, it raises several concerns. First, most papers claim to build technologies to solve communication barriers for deaf people, but many (12 of 43) datasets do not disclose who they collect data from. This indicates an underlying assumption: that everyone signs the same way or that variations in signing are insignificant. We unpack additional concerns below.

Second, even as datasets are diversified in terms of contributors, their usage is not. The three datasets that use interpreters only \cite{bobsl, phoenix} are long-standing benchmarks in the field, and are used by 41 of the 60 continuous sign language recognition works in our systematic review. All three of these datasets are continuous sign language and draw from existing media broadcasts. While these works offer large-scale annotated datasets to advance sign language recognition (which has known to be constrained by lack of data), the question arises whether is is appropriate to use interpreted datasets as source material to develop sign language AI. First, the majority of sign language interpreters are hearing users who may not sign in a manner that aligns with usage patterns in Deaf communities. Instances have been documented where Deaf viewers face challenges in understanding the interpreters in the same broadcasts used for ML purposes \cite{doi:10.1080/0907676X.2021.1936088}. Secondly, the nature of scripted and interpreted language use, especially under the constraints of simultaneous interpreting, diverges significantly from language in the wild. 
This may result in a distorted representation of sign languages in AI systems (see also \cite{SignOn_2022}). 
We note that authors of some of these datasets discuss limitations -- \eg BOBSL \cite{bobsl} remarks on ``translationese'' extensively -- but most works that use these datasets do not. These distortions have broader implications. Deaf end users may find themselves compelled to adjust their sign language use to accommodate the limitations of AI technologies trained on this data, a form of linguistic subordination to technology.

More recent datasets have recognized this gap between training data and target users, and sought to collect more representative data -- ASL Citizen \cite{desai2024asl} and Sem-Lex \cite{kezar2023sem} are both large scale isolated sign language recognition datasets of ASL, and aim to collect data from ``fluent'' DHH signers. While this is an improvement, details in how participants were recruited reveal that the notion of ``fluency'' is more subjective than what is discussed in either paper. ASL Citizen claims to recruit ``fluent signers'' from ``trusted groups'' but does not state what/who these are. In contrast, Sem-Lex defines ``fluent'' signers as those who acquired sign language in childhood. While people who acquired sign language in childhood are a portion of contemporary Deaf communities, it is not the only group, and not even the largest one. 95\% of deaf children are born to hearing families. Often these children do not learn sign language until later in life or at all, because medical practitioners often discourage parents from using a signed language \cite{murray2019education}. 
This illustrates the ideological meaning of ``fluency''. While later or different acquisition paths means they might sign differently from the ideological "norm", excluding them from datasets means we exclude them as users of designed technologies. While targeting subsets of the community can help scope data collection, our concern is how this bias is framed: Sem-Lex argues for data representative of ``deaf signers'' in general, without explicitly discussing how their data may not be representative of many signing Deaf people. Without this disclosure, we worry this may lead to applications that inadvertently marginalize a large proportion of Deaf communities. 

Overall, perhaps the biggest driver in mismatches between data and applications is the opposing goals of data as needed for machine learning applications and language as it happens in the world. First, finding an optima for machine learning necessitates scoping multi-dimensional and nuanced realities to something neat and tractable. Datasets make decisions about what variation is desirable to collect, and what is out-of-scope for a particular dataset. For example, ASL Citizen considers variation in background, illumination, and camera angle of recorded videos desirable, and Sem-Lex considers signer diversity across race and gender axes. At the same time, the prompting and labelling procedures in both datasets both seek to minimize label noise for signs for each category. In ASL Citizen, contributors are prompted to copy a seed signer's production of a sign, instead of providing their own sign for a concept. Similarly, in Sem-Lex, if a contributor provides a sign that is not included in a pre-defined corpus, it is discarded. This creates tension in the decision to collect a racially diverse dataset: even if Deaf people of color are represented, if a dataset only retains signs they produce that are present in dictionaries historically biased towards language used by white people \cite{https://doi.org/10.1111/lang.12540}, signs they use within their own communities may be discarded.

Clean data and high quality annotations are therefore in direct tension with procedures that foster agency and authenticity from signing contributors. This tension plays out in many different ML fields \cite{bender2018data}, but we are more concerned with how characterizations for desirable and excluded variation for datasets tie to a larger societal rhetoric of ``good'' and ``bad'' language. Revisiting our earlier discussion of fluency as an ideal, we note the concept of fluency is frequently entangled with notions of racial and ableist privilege, often being contingent upon closeness to whiteness and normative physical ability \cite{henner2023crip}. Without a critical examination of what constitutes ``fluency'', there is a risk of elevating those who, by virtue of early exposure to sign language and alignment with privileged identities (\eg racial, able-bodied), are considered the ``purest'' or most ``ideal'' users (also see `native' signer bias discussed in \cite{hochgesang2023w}). This paradigm risks overshadowing the diverse linguistic realities of deaf people and can again perpetuate a form of linguistic subordination to technology, where users are compelled to conform their signing to that of the ``ideal'', ``fluent'' model. This further overlooks the varied experiences of Deaf people with additional disabilities that might influence their interaction with sign language AI technologies, or even for Deaf people considered ``fluent" if they need to modify their signing (\eg they're signing one-handed because they're holding an object), in contravention with a goal of accessible design. 

But second, the need for large scale training data may engender reliance on more scalable data collection procedures \cite{bender2021dangers} (\eg collecting data from hearing interpreters, scraping from publicly available videos on the Internet, using subtitles) and result in suboptimal datasets that do not capture language as used by deaf people. We discuss this more in the next section, but for now, we ask the question: who gets to decide whether using or collecting more data outweighs the possibility that data may lead to biases that marginalize \cite{bender2021dangers}?

\subsection{Labels lack linguistic foundation}

Next, we looked at the annotation schemes used by models, which we found to be a good proxy for understanding how models use (or misuse) prior linguistic knowledge. We find that half of the papers (51) rely on glosses -- a written language representation of signed language intended to preserve original meaning and structure \cite{comrie2008leipzig} -- as either their main output or intermediary representation. Specifically, we find 30 papers that use glosses alone, with an additional 17 using glosses alongside spoken language translations, 4 using glosses alongside phonological features or other annotations. 

We find that sign language AI research has adopted the use of glosses without discernment, and without following best practices pioneered in linguistics \cite{hodge2022good}. Glossing conventions in linguistics are closely tied to projects: there is no singular gloss system, and gloss systems vary depending on the theoretical framework and questions of the research team. This similarly happens in sign language datasets, regardless of whether the gloss system is intentionally designed, or a consequence of data processing. For example, WLASL \cite{li2020word} (an ISLR dataset) merges gloss systems from different scraped online resources, and this leads to a final gloss system largely based on their English literal - in this gloss system, the sign for \textbf{PRESENT} meaning gift, and \textbf{PRESENT} meaning time are represented by the same gloss\footnote{PRESENT - gift - \url{handspeak.com/word/3783/}
PRESENT - time - \url{handspeak.com/word/2751/}}. This is distinguished from ASL Citizen and Sem-Lex, which use a gloss system from ASL-LEX \cite{sehyr2021asl}, which distinguishes signs by their semantics (\eg \textbf{BOW\_1} meaning hair ornament, and \textbf{BOW\_2} meaning archery are given distinct glosses\footnote{BOW\_1 - \url{asl-lex.org/visualization/?sign=bow_1} BOW\_2 - \url{asl-lex.org/visualization/?sign=bow_2}}). There are still other glossing systems that would be useful for and employed by linguists in some contexts (\eg those that make finer distinctions between phonological variants of the same sign), that we did not find represented in current sign language AI research. Critically, glosses cannot represent all linguistic phenomena in signing, \eg signs that point or depict, name signs, etc. Researchers often rely on internal or current practices for additional conventions.

Second, while glosses generally make source languages accessible to those in the field who may not be fluent in both languages, they do not stand alone as a complete representation, and lose meaning like any translation. In linguistic research, glosses often accompany the source language as to provide some access to meaning for those not fluent. Unfortunately, in sign language research, glosses are often used as the \textit{only} representation of signs, without any direct link to the source (be it video, photos or drawings), even when the issues with this representation are known -- a phenomenon called the ``tyranny of glossing'' \cite{hochgesang2019tyranny, hochgesang2022managing}. 

Here, we are concerned that the use of glosses in sign language AI research goes one step further, where many papers treat glosses as an actual translation, rather than a context-dependent representation. This is evidenced by several observations. First, virtually no paper describes the underlying design of the gloss system they are predicting. Without knowing what is being predicted, models lack usefulness for linguistic applications. Second, many papers build predictors on several independent datasets. We consider this to be predicting several independent, if correlated and not fully disclosed, tasks - \eg WLASL predicts the English word associated with a sign, whereas ASL Citizen predicts semantic categories of phonologically distinct signs. However, many of these papers claim these predictors are accomplishing sign language translation, effectively claiming these distinct and disparate gloss systems as complete representations of sign language. Third, for continuous sign language, the field often approaches sign language translation as a two-phase pipeline consisting of movement from sign2gloss and gloss2text. However, discussion is often not given to how the gloss system may bottleneck information (\eg if spatial and temporal components are represented).

Even works that do not use glossing may face the same issues. 11 papers do not specify what kind of annotation system they use, but attempt ISLR through a classification framework. 
The target here impacts task difficulty and the final application.  
We also find papers that use different systems -- 4 works use phonological features, and 5 use other notation systems like HamNoSys -- systems which are also specialized, noisy, and tied to specific theoretical perspectives on signs \cite{hochgesang2014using}. 
Our point is not that glosses are inherently bad, rather that they are partial and subjective representations of sign language and deeply shape the task at hand. 
When researchers focus on improving model performance without contextualizing what they are even predicting, they fail to engage with a core part of the research. 
ML scholars need to be explicit with their design choices and articulate trade-offs between systems.

We also note a growing trend of end-to-end translation, where works use spoken language translations as targets (18 works in our review).
This is largely motivated by the difficulty and expense in acquiring high quality annotations for sign language data. 
These works instead often rely on subtitles for supervision. 
While one might think this avoids the issues above, it adds other considerations.
First, there is no guarantee that the subtitles reflect the same content or order of content, for a number of reasons.
In simultaneous work, the captionist or interpreter may miss content; in translation, the interpreter may need to inject additional context depending on audience; and if captions are automated, biases from technology can be injected (\eg automated captioning struggles with technical terms and accents).
But even in situations where the subtitles reflect reliable translations, translation itself may not be perfect.
For example, the lyrics of a song used in a sign language music video are technically accurate, but will miss the expressive art of the signer.
Generalizing on this example, by relying heavily on spoken corpora, we limit ourselves only representations that align with spoken language conventions, paralleling issues raised in Section 4.1. 
Finally, that most work focuses on mapping sign languages to spoken languages (including glossing) is uncomfortable, because it echos misconceptions that sign languages are not independent, but analogues of spoken languages.

Overall, despite sign language modeling being framed as an computer vision \textit{and} natural language processing problem, we find there is a lack of linguistic awareness and incorporation of linguistic knowledge into research approaches. This leads to researchers appropriating annotation schemes without context (such as glossing), prioritizing ease rather than quality (such as subtitles), and over-relying on semantic representations (tied to spoken languages, rather than other representations that offer other applications).

\subsection{Modeling decisions inherit biases}

Next, we looked at machine learning modeling decisions.
Of the 101 papers in our review, we found that 59 models use vision-based inputs (\ie RGB video or images), 34 use pose-based inputs (\ie joint keypoints estimated from videos by a pose extractor), and 10 use other input representations (\eg manually assigned features or 3D sensor data). Note some works use multiple inputs. 

Data-driven AI-approaches typically rely on large amounts of annotated data to train.
As most sign language datasets are small, many works will employ transfer learning approaches, where sign language models will fine-tune or rely on outputs from previous models pretrained in another setting, where data is more abundant.
However, transfer learning is not without its risks: pretraining can introduce biases into models that are inherited by fine-tuned models \cite{wang2023overwriting}. 

From this perspective, it is concerning that 34 of the papers use pose-based inputs, which are extracted from pre-trained pose estimators \cite{lugaresi2019mediapipe, cao2017realtime, fang2022alphapose}. 
These are models not trained on sign language data, but using action or gesture videos.
Moryossef \etal \cite{moryossef2021evaluating} show failure models and biases when applying them to sign language: for example, handshapes in sign language are typically much more fine-grained than what these models encounter in pre-training.
Furthermore, by construction, many pose-based models exclude information necessary to understand sign language: for example, even though MediaPipe \cite{lugaresi2019mediapipe} extracts facial landmarks, Selvaraj \etal \cite{selvaraj2021openhands} advocate for the use of a reduced set of keypoints that include no information about facial expression, even in continuous sign language settings where the face is critical to grammar.

Similarly, many of the vision-based models (42 of 59 models) also employ pre-training. 
24 of these models only pre-train on non-sign language datasets (with ImageNet \cite{deng2009imagenet}, a natural image dataset, and Kinetics \cite{carreira2017quo}, a human action dataset being most common).
Again, it is unclear what biases are inherited with this approach: previous work by Desai \etal \cite{desai2024asl} shows that models pre-trained on Kinetics provide no capability to recognize isolated signs beyond random chance, and work by Shi \etal \cite{shi2022open} suggests that pre-training on ImageNet may in some cases, degrade performance.
While the other 18 models do explore pre-training on sign language datasets instead, the majority of these works pre-train on BSL (8 models) or ASL (8 models). 
These models often then evaluate on other sign languages, and although we consider this pre-training to be a closer domain than \eg action videos, it is unclear if this introduces any biases in phonology shared between the sign languages versus distinct.
In our analysis, we identified no paper that provided a quantitative analysis of potential biases from pre-training: even though as papers compare pre-training versus training from scratch \cite{jang2022signing} or different pre-training datasets \cite{shi2022open}, all papers report overall metrics on datasets exclusively, without seeking to understand if performance increases come with trade-offs (\eg reporting metrics class-by-class to understand if improving recognition of some signs comes at the expense of others). 

Beyond pre-training, a second sub-theme that we observed is that even as some papers claim to produce general methods, it is unclear if methods are correcting issues cascading from previous design decisions.
An interesting case example is in Zuo \etal \cite{zuo2023natural}, which argues that semantic similarity in English glosses can be used to improve sign language recognition, as sometimes signs related in meaning share phonology.
However, to demonstrate this claim, this paper relies primarily on internet-scraped datasets that rely upon English glosses to merge and distinguish signs, including WLASL \cite{li2020word}.
Our exploration of this dataset suggests that this procedure creates artifacts where distinct glosses refer to identical signs in ASL (\eg ``DORM'' and ``DORMITORY''), and it is unclear if improvement from the proposed method is due to correcting these artifacts versus general linguistic properties of sign languages.

Overall, we observe the majority of sign language AI works build off previous methods, with known issues and flaws in how they represent sign language.
While this point is understandable because re-inventing every design aspect of a new sign language model is unreasonable for any individual paper, this means that issues are inherited by future models, often uncritically. 
Echoing perspectives from previous sections, we argue that in many cases this is because authors lack the linguistic expertise to fully identify where modeling decisions not be representative or general. 
This creates systemic biases in modeling that align with decisions made due to convenience (\eg it's easier to use existing pose-based models, rather than training one specialized to sign language), but ultimately become standards as new papers do not re-assess if these design decisions align with sign language, but uncritically adopt them as defaults.

\section{Calls to Action}

Synthesizing our results, we take the position that as a field, sign language AI research lacks \textit{intentionality}: collectively, problem formulation and model design is 
not guided by what best aligns with Deaf stakeholder interests or growing trends in sign language research that center the complexities of lived deaf experiences. 
In the absence of these guiding principles, these decisions are left to researcher preference and ease. 
We showed that in spite of a range of possible problem formulations, datasets, targets, and models, most works narrow to a few defaults.
Although our point is that this is problematic even if every paper is well-executed, we expose numerous issues to demonstrate that these biases are likely induced by positionality, as most research is led and conducted by hearing non-signing researchers.
That most research is motivated by communication barriers is tied to the issue that many researchers view deaf people as being `deficient'. 
That most papers use datasets or prediction targets that misalign with broader Deaf languaging patterns connects to how many authors lack linguistic knowledge and actual engagement with Deaf communities. 
Some of these misaligned decisions are now baked-in as standards, such as the use of interpreter-only datasets as benchmarks, or the use of pretrained models without fully understanding their biases.
These misalignments have the potential to marginalize the very target users of sign language technologies. Moreover, as Deaf signing communities are a wide spectrum, they may marginalize subsets of the community even as they serve others. 

Towards addressing this systemic issue, we advocate that the field foster Deaf leadership.
Previous works have advocated for including Deaf collaborators \cite{yin2021including}, and while we agree that Deaf-hearing collaboration is essential to make meaningful progress in the field, we also believe that including Deaf people in each individual project is not a structural solution.
First, just including Deaf collaborators does not necessarily mean they are driving the research agenda. 
In most cases, they are not. 
In the first-hand experience of the authors of this paper, Deaf researchers are often only asked to collaborate often well after the idea has been conceived, the team built, the research conducted, or even near the project write-up as sometimes the sole "deaf" person.
In this paper, we showed that there are often tensions between how to allocate limited resources in projects and making decisions that are linguistically and culturally appropriate.
Currently, most of these decisions are made by hearing (often non-signing) researchers, and sometimes this is done even without awareness that an impactful decision is being made. 
``Leadership by the most impacted'' is one of the core principles of Disability Justice \cite{berne2018ten}
: even if Deaf researchers may not have all the answers in these complex trade-offs, enabling us to lead research means these decisions are at least being made by those with a larger stake.

But second, Deaf researchers are underrepresented in the field, and even if exclusionary structures are fully addressed, may still persist as a minority for demographic reasons.
Asking DHH scholars to be involved in each individual project creates burden given the overwhelming number of sign language AI works relative to the number of DHH researchers, and may distract them from other priorities or create tensions where they feel declining a project harms their community \cite{angelini}.
Instead, the field needs to contend with how to amplify Deaf perspectives, even as they may continue to form a minority of research outputs. 
Towards this end, hearing researchers should reassess their role in work involving Deaf signing communities.
Rather than being the ones to dictate the agenda and be the public face, hearing researchers can transition these opportunities to Deaf researchers, and instead switch to a role of supporting Deaf researchers like taking on the responsibility of accessibility or promoting their training. 

For this to be possible, all researchers in sign language AI research need to be transparent about their positionalities.
This imperative extends beyond a `confession before a crime', aspiring instead to weave positionality deeply into the research, enhancing transparency and underscoring the impact of researchers' backgrounds, experiences, privileges, and biases in their work. 
Transparency about one's positionalities is an increasingly recognized practice in sign language linguistics, sociolinguistics, interpreting studies, and Deaf Studies research \cite{hou2017negotiating, kusters2017innovations, kusters2022emergence, mellinger2020positionality, Hochgesang2022}.
where the lived experience of researchers (DHH or hearing) can significantly differ from those of their participants in aspects such as ethnicity, race, other disabilities, and educational and linguistic backgrounds.
Positionality statements are not a standard for sign language AI research (although some works informally disclose \cite{bragg2021fate, desai2024asl}), but given how the field contends with similar issues with potential mismatches between researchers and target users, we recommend it become adopted as practice. 

At the same time, we are cautious about our call for Deaf leadership. 
While we believe it is a meaningful step forward, it is not a full solution in itself, and followed uncritically, it risks corruption of the very principles we issue this recommendation under. 
We've noted that calls for and projects that claim Deaf collaboration or leadership have become tokenizing \cite{twitter_thread}.
We worry that our call for Deaf leadership may be similarly impacted. 
Without carefully considering whose voices to include, how to meaningfully build consensus, and how to reconcile disagreements, attention might focus on those who already have the most power, glossing over inequalities within the community.
Deaf researchers themselves must acknowledge there are gaps, and Deaf leadership must come from a wide range of perspectives and backgrounds.
We are careful to note our own positionalities (\eg educational and literacy privilege).
We further found critiques of our own work upon reflection (\eg ASL Citizen, which two authors on this paper worked on). 
Just because we are DHH doesn't mean we are immune to participating in systemic biases.

Thus, our call for Deaf leadership is intended to be a call for ongoing conversation, one in which we continuously re-evaluate how positionality influences research, and where stakeholders need to be in charge of decisions.
For example, even as we ask hearing researchers to transfer visibility and accountability to Deaf researchers, to what extent does this depend on the project, the discipline(s), and other people involved? 
And to which Deaf researchers?
Even now, these are questions we do not fully have the answers to. 
But to find answers, there first has to be a conversation taking place, which is currently absent from large swaths of the field. 
We invite all sign language AI researchers to join the conversation.

\section{Acknowledgements}
This work was supported by Center for Research and Education on Accessible Technology and Experiences (CREATE) and partially funded by a gift from Microsoft. We would like to thank Cassandra Kim for valuable help with annotation in our systematic analysis. We would also like to thank Danielle Bragg, Hal Daumé III, and Mary Gray for continued discussions that sparked ideas that informed this work. 

\section{Bibliographical References}\label{sec:reference}
\bibliographystyle{lrec-coling2024-natbib}
\bibliography{manuscript}

\appendix
\section{Methods Supplementary and Datasets}

We used an iterative process to develop questions for our systematic analysis, guided by an in-depth qualitative review of a few papers by all authors. Papers for this qualitative analysis were nominated by authors based upon individual authors' beliefs that they were representative of current modeling work, or would generate multidisciplinary discussion. Our final questions focused on four different themes: framing of the research in abstract or introduction, datasets used by the papers and other inputs for modeling, annotation or labeling schemes used for model outputs, and the use of pretrained models anywhere in the ML pipeline. Two annotators coded each paper, and a third annotator was called in to resolve disagreements. Two of the annotators have a background in ML and are familiar with reading such papers, one annotator has a background in psycholingusitics. 

Of the 101 papers in our review, we find that 60 work with continuous sign language datasets, 26 work with isolated sign language datasets, 3 with a combination of isolated and continuous sign language datasets, and 11 work with fingerspelling data (8 focus on recognition from images, 3 study fingerspelling in a continuous signing context aka in-the-wild). 

There are a total of 43 publicly available datasets used across our corpus (each used to varying degrees). Seven works collect their own private dataset. The sign languages studied in the public datasets include the following: American Sign Language (ASL), Deutsche Gebärdensprache (DGS), Chinese Sign Language (CSL), British Sign Language (BSL), Turkish Sign Language (TSL), Russian Sign Language (RSL), Indian Sign Language (ISL), Lengua de señas argentina (LSA), Greek Sign Language (GSL), Lengua de Signos Española (LSE), Arab Sign Language (ArSL), Bangla Sign Language (BdSL), Vlaamse Gebarentaal (VGT), along with some multilingual datasets (JWSign, SP-10). We note that along with disparities in who contributes data, not all sign languages are equally represented.

\end{document}